\documentclass[english]{article}
\usepackage[latin9]{inputenc}
\usepackage{amssymb}
\usepackage{babel}
\begin{document}

\title{The Computational Theory of Intelligence: Data Aggregation}

\author{Daniel Kovach}
\maketitle
\begin{abstract}
In this paper, we will expound upon the concepts proffered in \cite{cti},
where we proposed an information theoretic approach to intelligence
in the computational sense. We will examine data and meme aggregation,
and study the effect of limited resources on the resulting meme amplitudes.
\end{abstract}

\section{Introduction}

In the previous paper \cite{cti}, we laid the groundwork for what
was referred to as computational intelligence (CI). In this paper
we will further this discussion and offer some new insights. We discussed
how intelligence can be thought of as an entropy minimizing process.
This entropy minimization process, however comes at a cost.

Recall from \cite{cti,thermo} that whenever a system is taken from
a state of higher entropy to a state of lower entropy, there is always
some amount of energy involved in this transition, and an increase
in the entropy of the rest of the environment greater than or equal
to that of the entropy loss \cite{thermo}. The negative change in
entropy will require some amount of work, $\Delta E$.

The central concept of this paper concerns how smaller concepts aggregate
together into larger and more complex structures. This process is
a function of the complexity of each element, and the energy available
to the system. The complexity of the components will delegate the
degrees of freedom available to associate with parts of other constituents.
The amount of energy available must be enough to supply the constituents
with energy necessary to fulfill this task. We call this association
between participating degrees of freedom a \emph{bond}, and assert
that the degrees of freedom that form the bond are covariant and thus
have minimal entropy than that of the other degrees of freedom, respectively.

Though we will primarily be concerning ourselves with numerical data,
it should be stressed that these concepts apply to a panoply of possibilities
including everything from subatomic particles, to atoms, to social
groups, to politics. Here, the concept of \emph{energy} is taken in
the computational sense. That is, \emph{computational energy} refers
to the amount of computational work required to form bonds.

\section{Creating Structure}

In virtually all facets of nature, more complicated structures are
formed by less complicated structures, after the addition of some
amount of energy. Suppose the set $C$ contains $I$ constituents,
each with a certain quantity of degrees of freedom $J$ as expressed
in notation:$c_{ij}\in C,i\in\{0,..,I\},j\in\{0,...J\}$. For example,
a carbon atom has four valence electrons available for use in forming
bonds. Note that the $c_{ij}\in C$ need not be homogeneous, as (continuing
with our example from Chemistry) a carbon atom is free to form bonds
with other elements and molecules. We shall call the aggregation of
these constituents a \emph{meme}, a term coined by Richard Dawkins
in \emph{The Selfish Gene}, \cite{dawkins}.

Let us further suppose that with each of these degrees of freedom,
$j$, there is a certain amount of energy required to participate
in the formation of a more complicated structure, and all degrees
of freedom participate. This relationship may not be linear. Let $\rho_{E}(c_{ij})\mapsto\mathbb{R}$
denote the energy for each respective $c_{ij}$. Then the total energy,
$E$, required to form a more complex structure, or \emph{activation
energy}, from $c_{ij}\in C$ is therefore

\begin{equation}
E=\sum_{i}^{I}\sum_{j}^{J}\rho_{E}(c_{ij}),c_{ij}\in C
\end{equation}

Agan, it is worth noting that although the term 'energy' derives from
physics, we can take it in this particular context to have a more
general meaning, in terms of the computational effort required to
establish the bond between degrees of freedom.

\subsection{Applications to Data Mining}

We can begin to draw immediate relevance to data mining. To make sense
of a data set, we must determine the elements that are correlated
and thus have minimal entropy with respect to the others. Some of
these degrees of freedom will participate, others will not bear any
relevance. Further, some elements may not contain any information
at all, as is the case with sparse data sets.

It is an active area of research to sieve out data with a higher information
content for use in classification or clustering. Processes such as
PCA or Cholesky transformation are used to reduce the amount of feature
vectors for classification and find a transformation to a basis. This
has a clearly defined interpretation in Linear Algebra, though we
argue that information content may be another potential approach.

It is not uncommon to deal with data sets such that each element is
characterized by thousands or hundreds of thousands or feature vectors
or more, \cite{features}. Computationally intensive techniques like
PCA or the Cholesky transformation may prove intractible for data
sets of this magnitude.

Under our framework here, using entropic self organization \cite{cti}
the entropy contribution of redundant data would be very small, and
that of random data very large. In such a context, meaningful data
would be seen as a comparable deviation between these two extremes.

All the notions of this section assume, of course, that the data has
been properly preprocessed. After all, treating all data in an unfiltered
topological sense would fail for such data sets as the iris data set,
and in natural language processing.

\section{Meme Amplitude\label{sec:Meme-Amplitude}}

Now that we have determined the total energy necessary for a meme
to reach the next state, lets talk about the transition from one state
to another. In this case. Let $\triangle E=E_{2}-E_{1}$, or the difference
between the activation energy and the \emph{resting energy}, or the
energy required to sustain the elements unadulterated. Using $\triangle E$,
we can speak of the \emph{relative energy}, $E_{r}$ or the initial
energy less $\triangle E$. Let us further suppose that this $E_{r}$
can be described by some function, $y$ which we may also refer to
as the \emph{meme amplitude.} 

Let us take into account some considerations for $y'(t)$. The rate
of change in $y(t)$ is almost certainly proportional to $y(t)$ itself,
as the rate of change in the energy of the system should be proportional
to that which it already has. If we include a constant of proportionality
which we will call \emph{affinity} $A$ then we have the familiar
$y'=Ay$, but taking into account the fact that $\Delta E$ is a boundary
for $y$, we have the familiar logistic equation:

\begin{equation}
y'=\frac{A}{\triangle E}y(\triangle E-y)\label{logistic}
\end{equation}

onto which we imposse the following boundary conditions
\begin{enumerate}
\item At $t=0$, we start at our initial state, where $E_{r}=0$
\item As $t\rightarrow\infty$, $E_{r}\rightarrow\Delta E$.
\end{enumerate}
The latter implies that after some time, the transition to $\Delta E$
should be final, or that $E_{r}$ should come arbitrarily close to
$\Delta E$ as time progresses.

\subsection{Solutions}

Equation \ref{logistic} has been well studied, and its solutions
fall into a broad class of functions known as \emph{sigmoid functions}
known for their characteristic 'S' shape \cite{DE}, and who's applications
range from their use as cumulative distribution functions in probability
and statistics to activation functions in artificial neural networks.
Due to the first boundary condition, we restrict the solution to $t\geq0$,
which cuts off the left hand side of this 'S' shape, or translates
the entire curve such that the lower asymptote is suitably close to
$t=0$.

\subsubsection{Growth Rate}

Some comments should be made on the growth rate $A$. This constant
will determine the rate of change in the sigmoid curve. With increasing
$A$, the curve gets steeper and the time to reach the meme amplitude
decreases. We will use the term \emph{aggressive} to refer to large
$A$\emph{. }For drastically large $A$, the sigmoid curve approaches
a step function. $A$ will also depend on the energy applied to the
system, $E_{A}$, and will itself have its own boundary conditions,
as the dynamics described by \ref{logistic} will often break down
for $E_{A}\gg\triangle E$. For example Carbon forms graphite at (relatively)
low energies, but for higher energies, it forms diamonds. This is
a function of the amount of valence electrons used in the bonding
process (3 and 4 for graphite and diamond, respectively). A similar
analogy may be made for data and the computational effort we are willing
to put into the entropic self organization algorithm.

\subsubsection{Amplitude}

The meme amplitude is certainly not restricted to a single meme amplitude,
at least not in the general case. Once the transition $E_{r}\rightarrow\Delta E$
has been, we are free to apply the logic successively to determine
the next successive meme amplitude and so forth. We will call the
repeated demonstration of meme amplitude growth via \ref{logistic}
the \emph{hierarchical aggregation paradigm}.

\subsubsection{Extrema\label{sub:Extrema}}

Although the solutions to \ref{logistic} are monotonically increasing,
some subtle nuances in its curvature provide useful insights. Consider
the second derivative, which is easily obtainable from \ref{logistic}

\begin{equation}
y''=\left(\frac{A}{\triangle E}\right)^{2}y(\triangle E-2y)(\triangle E-y)\label{eq:rate of change}
\end{equation}

The second rate of change happens at the roots of this equation, which
are $y=0,\frac{1}{2}\triangle E,\triangle E$. The first and last
are clearly the asymptotes, when the energy levels off. It is the
middle root that is of interest. After all, if sustainable growth
is to occur, there must be a transition in the sign of the growth
rate. This happens at $y=\frac{1}{2}\triangle E$.

\subsection{Stability and Sustainability\label{sub:Stability-and-Sustainability}}

The reality of the dynamics described by \ref{logistic} are not always
so clement and predicable in reality. Although often the affinity
may be treated as constant to make calculations and estimations easier,
in reality it may be a function of the applied energy, and valid in
some kind of threshold. If the applied energy level is outside of
this threshold, the model will break down.

Further, once the affinity has been determined it can be used as an
indicator of the \emph{stability}, that the transition to the next
energy level will occur. By observing the disparities between the
model \ref{logistic}, and observed data, we can gain an idea as to
if this transition will occur, or if the model will return to its
previous energy state, or even one below.

If we remove the boundedness condition, equation \ref{logistic} simply
becomes

\begin{equation}
y'=Ay\label{logistic-1}
\end{equation}

whose solutions are exponential for positive$A$. However unbounded
growth is unsustainable and leads to unpredictable behavior. If we
observe the behavior described in \ref{logistic-1} as opposed to
\ref{logistic}, we may have an indicator of collapse. Additionally,
if the rate of growth is not commensurate with the particular value
of $A$, the fecundibility of the energy level transition may be suspect. 

Once the transition to the next energy level has been made, there
is a certain amount of energy necessary to sustain it. If this energy
gets outside of the threshold, the model will not be able to maintain
this energy level. Thus the new energy level is not guaranteed to
indefinitely stable.

The calculation of $A$, its threshold, and energy to sustain it is
always contingent upon underlying factors unique to the system. These
must be taken into account, which may not be tractable, but perhaps
at least estimable.

\subsection{Bubbles}

Our discussion in section \ref{sub:Stability-and-Sustainability}
has many applications and is of great importance especially to society
as of the past five years. We can draw examples examples in real estate,
financial markets, and population growth. Indeed this mathematics
may be used to identify sustainable growth, or bubbles, periods of
extraordinary inflationary growth followed by abject collapse.

For example, consider the well studied behavior exhibited in the growth
of bacterial colonies. There is an initial lag phase as the RNA of
the bacteria start to copy, followed by an explosion of exponential
growth called the growth phase, follwed by a brief period of stability
before the bacteria exhaust their resources resulting in an expedient
decay.

\subsubsection{Determining Bubbles}

It may be difficult to determine whether a given data set is a bubble
in the making or the initial stages of a stable transition, though
we may gain insight by the application of two observations discussed
in this paper.

First, if we are able to calculate the affinity constant, and activation
energy, and we notice a disparity in the dynamics the system is exhibiting
and that which was forecasted using these constants, then we may have
reason to believe a bubble may be forming. Bubbles are caused by disproportionately
excessive growth or extraneous factors distorting the dynamics of
the system. But such factors are not perpetual and eventually the
true dynamics of the system will take over. At this point, the system
will fall to its natural energy state, commensurate with our calculations
and what the system is able to support.

Second, consider the rate of change. After all, we may not be privy
to the subtle inner workings of the system so as to calculate the
necessary parameters. This may be especially true in very large and
complicated systems, where subtle factors may contribute greatly to
the outcome of the calculation. We can fall back on our discussions
in \ref{sub:Extrema}. After all, if the system is exhibiting extraordinary
growth, in order to fit model \ref{logistic}, there must be a point
at which rate of change of the growth rate reaches zero as given in
\ref{eq:rate of change}. We may not be able to determine this a priori,
but we might be able to hypothesize whether the activation energy
inferred from a given point is fungible or not.

\subsubsection{Outliers}

The difficulty of calculating the activation energy of large and complicated
systems should be stressed. After all, there may be a panoply of factors
that will influence the overall dynamics of the model, and they can
be highly nonlinear.

\section{Competing Memes}

In the previous section, we considered the amplitude of only single
meme. Now we will consider meme amplitude in the presence of other
memes. In order to repeat the logic of section \ref{sec:Meme-Amplitude},
we must introduce the\emph{ interaction matrix} $\alpha_{ij}$, which
represents the effect meme $i$ has on meme $j$. The underlying mechanics
behind these $\alpha_{ij}$'s include resource allocation, and direct
competition. Thus, in the presence of $N$ memes, \ref{logistic}
becomes

\begin{equation}
\frac{dy_{i}}{dt}=\frac{A_{i}}{\triangle E_{i}}y_{i}(\triangle E_{i}-\sum_{j}^{N}\alpha_{ij}y_{j})\label{competition}
\end{equation}

where the constant$\triangle E_{i}$ can be absorbed into the interaction
matrix without a functional effect on the resulting solutions \cite{c1}.
Therefore \ref{competition} becomes

\begin{equation}
\frac{dy_{i}}{dt}=A_{i}y_{i}(1-\sum_{j}^{N}\alpha_{ij}y_{j})
\end{equation}

which is recognized as a Lotka Volterra equation commonly referred
to as the ``competition equation''. This is another very well studied
class of differential equations for which results can be found in
\textbf{\cite{c1,c2,c3,c4,c5,c6,c7}} just to name a few.

\section{Conclusion}

In this paper , we discussed some ramifications of the original principal
detailed in \cite{cti}. In essence, the subject of this paper could
be summarized as how data aggregates together to form more complex
memes, how this is effected by the presence of mutiple memes, and
the conditions under which this transition may or may not be stable.
We also looked at some practical examples of these principals in source
code.

There is still a great deal of room for future improvements. First,
we can improve our understanding of the affinity parameter $A$ and
how to better calculate it. We can also apply this knowledge to more
areas of study so as to be able to effectively calculate predictable
amplitude transitions, burgeoning growth or bubbles.

Additionally, we also wish to study the most basal layers of meme
aggregation and creation. The fundamental constituents, how they emerge
to form systems, and how these systems can aggregate hierarchically
as meme amplitude growth describes. We will also look at how this
process is perpetuated in computational processes, as opposed to information
in static elements and datasets.

\end{document}